\documentclass[journal]{IEEEtran}

\ifCLASSINFOpdf
\else
   \usepackage[dvips]{graphicx}
\fi
\usepackage{url}

\usepackage{graphicx}
\usepackage{amssymb}
\usepackage{cite}
\usepackage{underscore}
\usepackage[numbers,sort&compress]{natbib}
\usepackage{amsthm,amsmath,amssymb}
\usepackage{mathrsfs}
\usepackage{algorithm}
\usepackage{algorithmic}
\usepackage[colorlinks,linkcolor=red]{hyperref}
\usepackage{makecell}
\usepackage{multirow}
\usepackage{diagbox}
\usepackage{threeparttable}

\makeatletter

\newcommand{\Rmnum}[1]{\expandafter\@slowromancap\romannumeral #1@}
\makeatother

\begin{document}

\title{YouTube-GDD: A challenging gun detection dataset with rich contextual information}

\author{Yongxiang Gu, \IEEEmembership{Student Member, IEEE}, Xingbin Liao and Xiaolin Qin
\thanks{This work was supported in part by Sichuan Science and Technology Program (2019ZDZX0006, 2020YFG0010), the West Light Foundation of Chinese Academy of Sciences, National Academy of Science Alliance Collaborative Program (Chengdu Branch of Chinese Academy of Sciences - Chongqing Academy of Science and Technology) and Science and Technology Service Network Initiative (KFJ-STS-QYZD-2021-21-001).
{\em (Corresponding author: Xiaolin Qin.)}
}
\thanks{The authors are with Chengdu Institute of Computer Applications, Chinese Academy of Sciences, China and University of Chinese Academy of Sciences, Beijing, China (e-mail: guyongxiang19@mails.ucas.ac.cn; liao\_xingbin@163.com; qinxl2001@126.com).}
}

\markboth{Journal of \LaTeX\ Class Files, Vol. x, No. x, March 2022}
{Shell \MakeLowercase{\textit{et al.}}: Bare Demo of IEEEtran.cls for IEEE Journals}
\maketitle

\begin{abstract}
An automatic gun detection system can detect potential gun-related violence at an early stage that is of paramount importance for citizens security. In the whole system, object detection algorithm is the key to perceive the environment so that the system can detect dangerous objects such as pistols and rifles. However, mainstream deep learning-based object detection algorithms depend heavily on large-scale high-quality annotated samples, and the existing gun datasets are characterized by low resolution, little contextual information and little data volume. To promote the development of security, this work presents a new challenging dataset called YouTube Gun Detection Dataset (YouTube-GDD). Our dataset is collected from 343 high-definition YouTube videos and contains 5000 well-chosen images, in which 16064 instances of gun and 9046 instances of person are annotated. Compared to other datasets, YouTube-GDD is "dynamic", containing rich contextual information and recording shape changes of the gun during shooting. To build a baseline for gun detection, we evaluate YOLOv5 on YouTube-GDD and analyze the influence of additional related annotated information on gun detection. YouTube-GDD and subsequent updates will be released at \url{https://github.com/UCAS-GYX/YouTube-GDD}.
\end{abstract}


\IEEEpeerreviewmaketitle

\section{Introduction}
\IEEEPARstart{W}{ith} the continuous development of artificial intelligence, various AI technologies have been applied to an endless stream, among which object detection technology has been widely used in the fields of public security, autonomous driving, virtual reality and so on\cite{Zou2019}. As an important component in the field of security, video surveillance has gradually developed from traditional manual detection to automation and intelligence in view of the low efficiency and accuracy of manual inspection of dangerous guns in the current surveillance videos\cite{Elharrouss2021}.

Compared to general object detection tasks, gun detection is more challenging since the person holding the gun is always trying to hide it completely or partially. With recent advances in deep learning, this task is expected to succeed\cite{Shah2021}. In the past decade, convolutional neural networks (CNN) and general deep learning models have been widely used in automatic object detection. These models do not need to design features by hand, but learn features when training models on large-scale samples. On the whole, CNN-based detectors achieved the long-term state-of-the-art performance, while emerging Transformer-based detectors show impressive performance and great potentiality of application.

Anyhow, it is generally accepted that the quality of training samples imposes the performance upper bound of deep learning-based methods\cite{Brownlee2018,Shorten2019}. Unfortunately, according to our investigation, while object detection algorithms have advanced dramatically in the past few years, the development of public available gun datasets has not kept pace. Specifically, images of gun datasets are often collected through image retrieval of search engines. These samples are characterized by low image quality and the guns are often shown in close-up view. It means that existing data can provide limited supervision information since they are easy to recognize. What’s more, the distribution of existing datasets is far away from the distribution of surveillance videos, causing the limited application prospect.

To promote the development of security, this work presents a new challenging dataset called YouTube Gun Detection Dataset (YouTube-GDD). Apart from annotating the instances of gun, instances of person are also annotated to provide richer supervision information. Compared to other datasets, YouTube-GDD is "dynamic", containing rich contextual information and recording shape changes of the gun during shooting.

The remainder of this paper is organized as follows. Section 2 gives a brief analysis of the related work. Section 3 introduces the process of creating YouTube-GDD and shows the statistic analysis. Section 4 gives the experimental results of YOLOv5 on YouTube-GDD and finally Section 5 conclusions.
\section{Related Work}
\subsection{Modern Object Detection Systems}
Look back to modern object detection systems, its frameworks can be mainly divided into two categories: two-stage and one-stage. The former is represented by Faster RCNN\cite{Ren2016} and the latter by YOLO\cite{Redmon2016}. The major difference between two frameworks in structure is that the two-stage uses the sub network of region proposals (RPN) to assist in generating proposals, while the one-stage generates proposals directly on the feature map. A modern detector is usually composed of following parts: a backbone which outputs the feature map of the whole image, a neck or named FPN\cite{Lin2017} which fuses the feature maps of different scales to obtain multi-scale features (optional), and a head which is used to predict classes and bounding boxes of objects based on the proposals. In practice, a backbone is commonly pre-trained on ImageNet and a neck is usually composed of several bottom-up paths and several top-down paths\cite{Gu2021}.

In both two-stage and one-stage object detection algorithms, anchors are introduced in advance to roughly surround objects with preset anchors with different scales, length-width ratios and positions. Recently, there emerged some anchor-free object detection methods\cite{Duan2019,Law2018}. This kind of method does not locate the object by fine-tuning the anchor, but directly regresses the key points of the object, training and predicting the position and size of the bounding box. Compared with one-stage detection framework, the object detection framework based on key points has almost no disadvantage in detection speed, but can achieve good detection performance.

Transformer is one of the most impressive technologies of recent years. It was originally developed in natural language processing\cite{Vaswani2017} and has recently been applied in computer vision.  In object detection, Transformer is used in feature extraction networks to predict precise bounding boxes and their corresponding categories by establishing a wider range of dependencies with self-attention. To the best of our knowledge, DETR\cite{Carion2020a} was the first to use the standard encoder-decoder architecture of Transformer for the object detection task, while Swin Transformer v2\cite{Liu2021} built a full Transformer backbone and achieved the state-of-the-art performance on COCO\cite{Lin2014}.
\subsection{Gun Detection Algorithms}
The detection of such dangerous objects as guns has always been one of the important research subjects in the field of security. Roberto\cite{Olmos2018} introduced Faster RCNN for automatic pistol detection and achieved impressive results on the proposed dataset, providing zero false positives, 100\% recall. Furthermore, Rana\cite{Alaqil2020} chose MobileNetV2\cite{Sandler2018} as the feature extractor in Faster RCNN in order to adapt to mobile gun detection applications. Abdul\cite{cmc.2022.018785} took the advantage of the latest models such as YOLOv5s to achieve effective results and speed. What's more, pre-processing technique of blurring the background with gaussian blur was used to improve F1-score. Apart from using the graphical appearance of the weapon, Alberto\cite{VelascoMata2021} leveraged the extra human pose information to improve performance of the handgun detector. Specifically, heatmap-like images that represent pose are integrated with grayscale images outputted by the basic handgun detector. Review these works, current gun detection algorithms are still mainly based on the general object detection algorithm. The improvement approach of specific scenarios requires in-depth analysis of samples, we hope a new challenging dataset can inspire innovations.
\subsection{Gun Detection Dataset}
There have been some works on gun datasets. Pistol detection dataset\cite{Olmos2018} contains 3000 images of short guns with rich context in the background. The images selected from the internet contain one or more handguns in diverse situations including video surveillance contexts. Sohas weapon detection dataset\cite{PerezHernandez2020} contains 3255 images, which is formed by weapons and small objects that are handled in a similar way. It includes six different objects such as pistols, knifes, bills, purses, smartphones and cards. With the increasing resolution of video surveillance, high-resolution samples are required. Considering that gathering real data is tough and time-consuming, annotations and images of Synthetic Gun Detection Dataset\cite{Rub1} are generated within a 3D environment and animated using 3D game engines. It is the largest open synthetic gun detection dataset, containing 25,000 images of indoor scenes and 10,000 images of outdoor scenes. Although the number of Synthetic Gun Detection Dataset is large, there is inevitably a distribution gap between the real world.

Based on these considerations, we turned our attention to gun-related videos on YouTube. With the rise of video platforms in recent years, people are willing to share their life and fun through videos, including shooting records and gun-related popular science introductions. Thanks to the temporal nature of the video medium, gun-related videos can record different views of the same gun and the shape changes of gun during shooting. It means that these annotated images can provide rich contextual information and structure information of guns.
\section{YouTube-GDD}
\subsection{Image Collection}
YouTube is adopted as the only source to collect gun-related images. Firstly, we collected about 500 videos by searching for "gun", "shooting", "weapons show", etc., and prioritized 720p or the highest resolution available for download. Then, frames are splitted at twice the current frame rate for each video, i.e., if the current video frame rate is 30, frames 60, 120, 180 and so on will be captured. For adjacent frames, some static and repetitive fragments are filtered out by calculating the difference of pixel values and setting the threshold. Finally, we manually deleted all images about shooting games. All images are saved in JPG format and the name format of each image and corresponding label is set as "YouTube id\_original frame rate\_split frame rate\_ID". 
\subsection{Image Annotation}
LabelImg is used to annotate instances of gun and person in YOLO format\cite{Redmon2016}. All guns, including handguns, rifles, machine guns, etc., are uniformly labeled as “gun”. The annotation criterion follows the coarse-to-fine rule. At the first stage, workers are required to label four categories i.e, "person","gun","uncertain person" and "uncertain gun". When annotating, workers are required to merely rely on image information instead of video information. At the second stage, uncertain objects at the first stage will be judged by experts. Finally, YouTube-GDD contains two categories, namely "person" and "gun", corresponding to category ids 0 and 1, respectively.

\subsection{Dataset Statistics}
\begin{figure}[]
	\centerline{\includegraphics[width=\columnwidth]{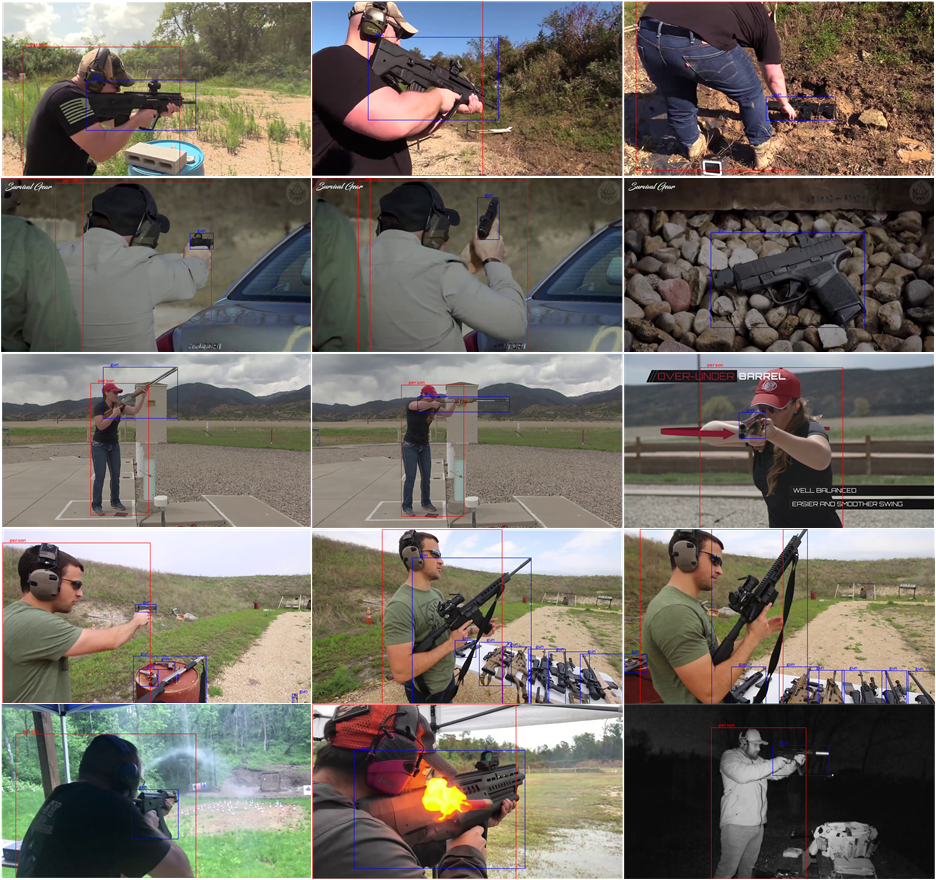}}
	\caption{Samples of annotated images in the YouTube-GDD dataset. Examples for each of the first four lines are captured from the same video while examples of the last line are captured from different videos.}
	\label{fig}
\end{figure}

On the whole, our dataset is collected from 343 high-definition YouTube videos and contains 5000 well-chosen images, in which 16064 instances of gun and 9046 instances of person are annotated. Examples of YouTube-GDD are shown in Fig. \ref{fig}
After annotating, we need to divide the entire dataset into training set, validation set and test set. Since the general division ratio is 8:1:1, we split the entire dataset into 10 non-overlapping folds by filename, each containing 500 images. The following indicators are concerned in YouTube-GDD: (1) The number of person and gun instances; (2) The number of objects with different scales; (3) How many videos does each fold came from. Following the standard of COCO, an object with pixels less than $32\times32$ is defined as a small target, an object with pixels between $32\times32$ and $96\times96$ is defined as a medium object, an object with pixels more than $96\times96$ is defined as a large object. Detailed statistical data are shown in Tab. \ref{table1}. It should be noted that source video files of adjacent folds may have one video overlap.
\begin{table}[H]\centering
	\caption{Statistic Information of YouTube-GDD}
	\label{table1}
	\small
	\setlength{\tabcolsep}{3pt}
	\renewcommand\tabcolsep{6.0pt}
	\begin{tabular}{ccccccc}
		\hline
		\multirow{2}{*}{\makecell[c]{{Split}}} &
		\multicolumn{ 2}{c}{{Instance}}& 
		\multicolumn{ 3}{c}{{Scale}} &
		\multirow{2}{*}{\makecell[c]{{Videos}}} \\
		\cline{2-6}
		& \makecell[c]{person} &\makecell[c]{gun} &\makecell[c]{small} &\makecell[c]{middle} &\makecell[c]{large} & \\
		\hline
		fold1 &467&1265&373&235&1124&35  \\
		fold2 &430&620&4&84&962&34  \\
		fold3 &466&905&39&259&1073&31  \\
		fold4 &427&751&5&124&1049&31  \\
		fold5 &471&716&11&120&1056&36  \\
		fold6 &415&718&13&122&998&43  \\
		fold7 &394&879&67&193&1013&42  \\
		fold8 &475&636&1&60&1050&34  \\
		fold9 &460&589&3&57&989&33  \\
		fold10 &518&953&37&281&1153&32  \\
		\hline
		all &9046&16064&1106&3070&20934&343  \\
		\hline
	\end{tabular}
\end{table}
Specifically, we firstly calculate the ratio of the different scales of bounding box in the entire dataset as the probability distribution, and then calculate the scale distribution of the bounding box of each fold. The two folds with the lowest JS divergence are chosen as test set and validation set, i.e, fold7 is chosen as the test set and fold6 is chosen as the validation set while the rest are chosen as the training set.
\section{Experiment}
\subsection{Experiment Setting}
All experiments are implemented based on the official YOLOv5 v6.0 project\cite{yolov5}, using YOLOv5s as the basic configuration. All models are trained on a NVIDIA RTX3090 (24GB memory) and Pytorch version 1.9.0. COCO\cite{Lin2014} is used to pre-train the backbone and feature pyramid network.  

Consistent with the default settings of YOLOv5 project, the input image size is set as 640×640, initial learning rate as 0.01, momentum as 0.937, attenuation coefficient as 0.0005 and batch size as 16 during the experiment. Stochastic Gradient Descent (SGD) optimizer is adopted and training Epochs are set to 300. During the training, the first three Epochs are preheated by warmup algorithm. After predicting, Non-Maximum Suppression (NMS) algorithm is used for post-processing.  
\subsection{Evaluation Metric}
For evaluation, we adopt the standard COCO-style evaluation metrics, i.e., $AP$ and $AP_{50}$. $AP$ is evaluated under different IoU thresholds, ranging from 0.5 to 0.95 with an interval of 0.05. In addition, the number of parameters and floating point operations are also counted.
\subsection{Result Analysis}

\begin{table}[h]\centering
	\caption{Experiment Results on YouTube-GDD Test Set}
	\label{table2}
	\small
	\setlength{\tabcolsep}{3pt}
	\renewcommand\tabcolsep{2pt}
	\begin{threeparttable}
	\begin{tabular}{ccccccccc}
		\hline
\multicolumn{1}{c}{\multirow{2}{*}{Method}} & \multicolumn{1}{c}{\multirow{2}{*}{w/ TL}} & \multicolumn{1}{c}{\multirow{2}{*}{w/ AoP}} & \multicolumn{1}{c}{\multirow{2}{*}{GFLOPs}} & \multicolumn{1}{c}{\multirow{2}{*}{Params}} & \multicolumn{2}{c}{Gun} & \multicolumn{2}{c}{Person} \\
\cline{6-9}
\multicolumn{1}{c}{}                        & \multicolumn{1}{c}{}                         & \multicolumn{1}{c}{}                         & \multicolumn{1}{c}{}                        & \multicolumn{1}{c}{}                        & $AP_{50}$         & $AP$        & $AP_{50}$           & $AP$         \\
		\hline
		\multirow{4}{*}{YOLOv5s} &&&15.80&7.01M&67.7&41.0&-&-      \\
		&         &$\checkmark$       & 15.81  & 7.02M  & 67.9 & 41.3 & 90.3 & 75.0 \\
		& $\checkmark$       &         & 15.80  & 7.01M  & 75.0 & 52.0 &-      &-      \\
		& $\checkmark$       & $\checkmark$     & 15.81  & 7.02M  & 77.3 & 52.1 & 92.4 & 81.2 \\
		\hline
	\end{tabular}
 \begin{tablenotes}
	\footnotesize
	\item $^*$TL means Transfer Learning and AoP means Annotations of Person.
\end{tablenotes}
\end{threeparttable}
\end{table}
As shown in Tab. \ref{table2}, we evaluate YOLOv5s on YouTube-GDD and report its performance on test set. When training from scratch, YOLOv5s trained with additional annotations of person can gain 0.2 and 0.3 points improvement on $AP_{50}$ and $AP$. With transfer learning, YOLOv5s trained with additional annotations of person can gain 2.3 and 0.1 points improvement on $AP_{50}$ and $AP$ even though the initial weights were trained on COCO which already contains annotations of person. Specifically, person is the subject that uses guns in most cases. Therefore, annotations of person could naturally provide additional information that could help identify the gun. What's more, this strategy needs negligible extra cost. Firstly, GFLOPs and parameters increase a little when adding the person classification branch in the detection head of the pure gun detection model. Secondly, there have existed well-trained person detection models, which can provide pseudo annotations for gun detection.

\section{Conclusion}
We collect and publish a new challenging gun detection dataset called YouTube-GDD. All images are captured from high-definition Youtube videos, leading to the high quality of training samples. Then, training set, verification set and test set are naturally divided according to different video sets, posing a challenge to the robustness of existing object detection algorithms. What's more, we introduce annotations of person into gun detection, discovering that additional related annotated information can promote gun detection performance.

In the future, we will continue to expand YouTube-GDD with high-quality annotated samples and optimize existing gun detection algorithms based on gun detection characteristics. 

\section*{Acknowledgement}
We thank Lab students, namely Jingyang Shan, Siqi Zhang, Yuncong Peng, Qianlei Wang, Gang Luo, Xin Lan, Boyi Fu, Yangge Qian, for their suggestions about improving the YouTube-GDD dataset.
\bibliographystyle{IEEEtran}
\bibliography{ref}

\end{document}